\documentclass[
]{ceurart}

\usepackage{listings}
\usepackage{adjustbox}
\usepackage{subfigure}
\begin{document}

\copyrightyear{2022}
\copyrightclause{Copyright for this paper by its authors.
  Use permitted under Creative Commons License Attribution 4.0
  International (CC BY 4.0).}

\conference{Woodstock'22: Symposium on the irreproducible science,
  June 07--11, 2022, Woodstock, NY}

\title{INO at Factify 2: Structure Coherence based Multi-Modal Fact Verification}


\author[1]{Yinuo Zhang}[%
orcid=0000-0001-7910-7983,
email=catrinbaze@gamil.com
]

\author[1]{Zhulin Tao}[%
orcid=0000-0001-9011-8464,
email=taozhulin@gmail.com
]

\author[2]{Xi Wang}[%
orcid=0000-0002-6642-8160,
email=wangxiboss@163.com,
]
\corref{cor1}

\author[1]{Tongyue Wang}[%
orcid=0000-0003-1897-4420,
email=apri.sunmoon@gmail.com,
]

\address[1]{Communication University of China, Beijing, China}
\address[2]{Institute of Information Engineering, Chinese Academy of Sciences, Beijing, China}

\cortext[cor1]{Corresponding author.}

\begin{abstract}
This paper describes our approach to the multi-modal fact verification (FACTIFY) challenge at AAAI2023. In recent years, with the widespread use of social media, fake news can spread rapidly and negatively impact social security. Automatic claim verification becomes more and more crucial to combat fake news. In fact verification involving multiple modal data, there should be a structural coherence between claim and document. Therefore, we proposed a structure coherence-based multi-modal fact verification scheme to classify fake news. Our structure coherence includes the following four aspects: sentence length, vocabulary similarity, semantic similarity, and image similarity. Specifically, CLIP and Sentence BERT are combined to extract text features, and ResNet50 is used to extract image features. In addition, we also extract the length of the text as well as the lexical similarity. Then the features were concatenated and passed through the random forest classifier. Finally, our weighted average F1 score has reached 0.8079, achieving 2nd place in FACTIFY2. The code is available at \url{https://github.com/Catrin-baze/INO-of-factify}.

\end{abstract}


\maketitle

\section{Introduction}

In recent years, the number of fake news has exploded due to the wide application of social media and the development of new technologies such as Deepfake. For example, research \cite{allcott2017social} indicates that fake information appeared in the three months before the election, which favoring Trump shared a total of 30 million times on Facebook, while those favoring Clinton were shared 8 million times. The rapid development of social media has led to the widespread dissemination of fake news, which not only affects people's lives, causes public panic, disrupts social order, affects public opinion, and manipulates the focus of the public but also damages the credibility of social media platforms. In 2018, the article "The Science of Fake News" \cite{lazer2018science} published in Science stated that falsehood diffused significantly further, faster, deeper, and more broadly than the truth in all categories of information. Therefore, effectively detecting fake news on social media to suppress the spread of fake news is of great significance for maintaining social stability and cyberspace security.

In order to meet the challenges of fake news, artificial intelligence, and deep learning technology are likely to play an essential role. Since 2017, the U.S. Defense Advanced Research Projects Agency(DARPA) has held a "Media Forensics Challenge" \cite{jin2020media}competition to promote misinformation detection technology developed rapidly. It is worth noting that the fragmented digital media environment provides a breeding ground for the uncontrolled dissemination of misinformation. Multi-modal fake news that combines graphics and text is more difficult to identify, posing a severe challenge for automatic multi-modal fake news detection. At the same time, combining multiple modalities has also been applied in various fields, such as recommendation systems\cite{SLMRec}\cite{EliMRec}. Many representative methods\cite{singhal2020spotfake+}\cite{zhou2020safe}\cite{jin2017multimodal} and datasets have been proposed to address the multi-modal fake news detection problem. Using multi-modal information to detect fake news has many advantages as different modalities capture different dimensions of the news article, and they can complement each other while evaluating the genuineness of the article.

In our paper, Our core idea is to thoroughly compare the correlation between claim and document from multiple perspectives as a basis for classification. Therefore, we propose a structure coherence-based Multi-Modal Fact Verification. We thoroughly compare the consistency of the claim and document from structure coherence, which contains four aspects: sentence length, vocabulary similarity, semantic similarity, and image similarity. Experiments prove the effectiveness of our method.

In the rest of this paper, we organize the content as follows. Data description and task definition are introduced in Section 2. Section 3 introduces related work. Our approach is described in Section 4, and experiments with results are discussed in Section 5. The conclusion of our work is presented at the end of the paper.



\section{The Task and Dataset}
\subsection{Task definition}
\begin{figure}[htbp]
  \centering
  \subfigure[]{\includegraphics[width=0.45\textwidth]{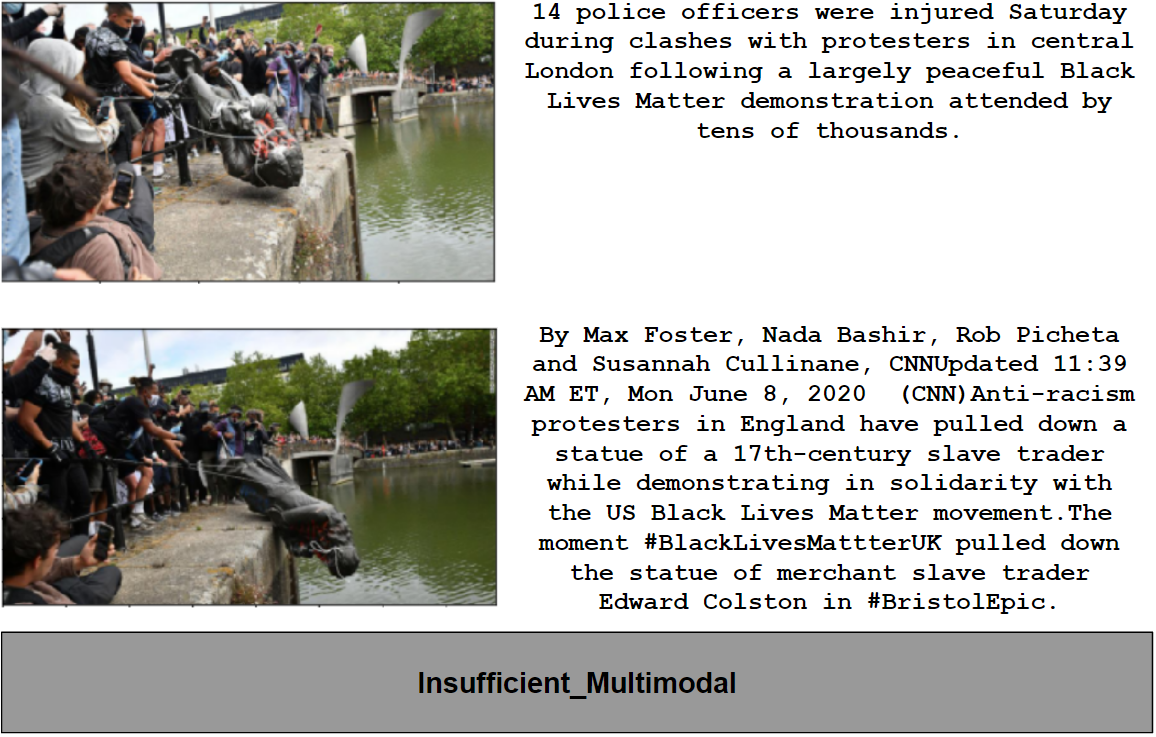}}
  \subfigure[]{\includegraphics[width=0.45\textwidth]{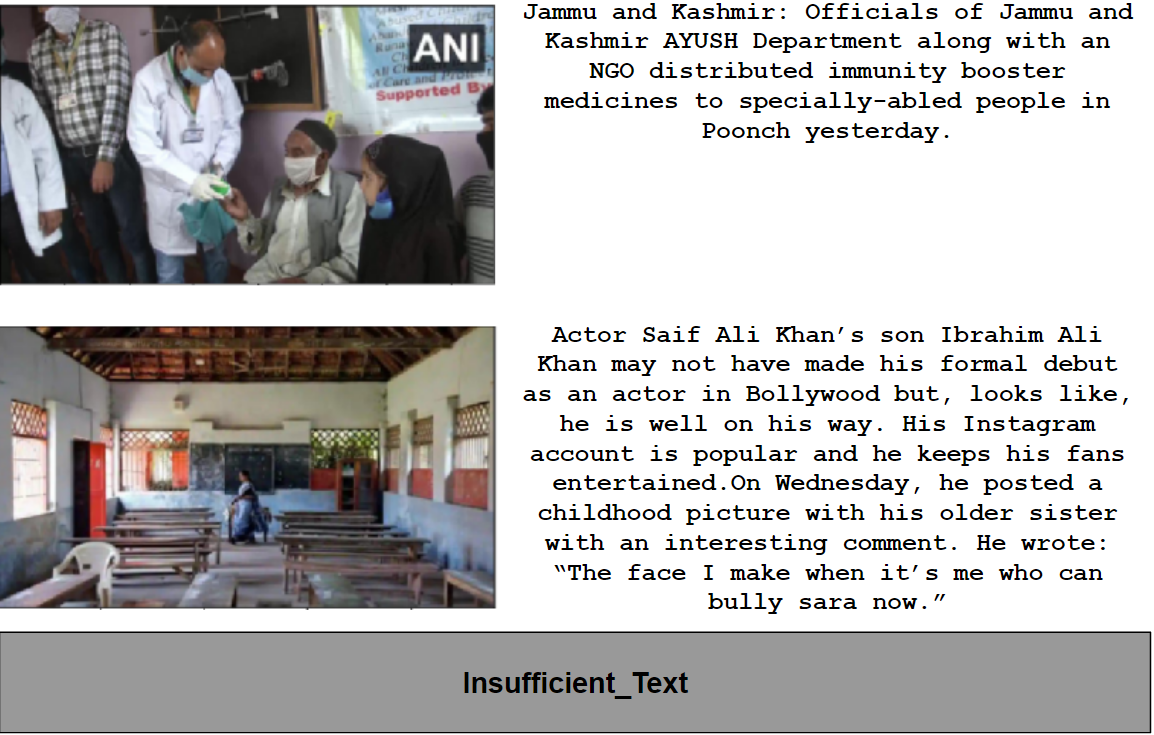}}
  \caption{An example of data.}
  \label{fig:1}
\end{figure}

The multimodal fact verification task FACTIFY2\cite{surya2023factify2}\cite{surya2023factifyoverview} is essentially modeled as a multimodal entailment. Each data point contains a claim to be detected and a reliable source document. Both of them are multimodal and contain an image-text pair. Therefore, the task is to give a textual claim, claim image, document, and document
image. The system has to classify the data sample into one of the five categories. The descriptions of the labels are as follows.
\begin{itemize}
\item \verb|Support_Text| : the claim text is similar or entailed but images of the document and claim are not similar;
\item \verb|Support_Multimodal| : both the claim text and image are similar to that of the document;
\item \verb|Insufficient_Text| : both text and images of the claim are neither supported nor refuted by the document, although it is possible that the text claim has common words with the document text;
\item \verb|Insufficient_Multimodal| : the claim text is neither supported nor refuted by the document but images are similar to the document;
\item \verb|Refute| : The images and/or text from the claim and document are completely contradictory i.e, the claim is false/fake.
\end{itemize}
Figure\ref{fig:1} shows an example of the data at FACTIFY, the data form of FACTIFY2 is the same as it.
\subsection{Data description}
The data collected from Twitter handles of Indian and US news sources: Hindustan Times\footnote{\url{https://twitter.com/htTweets}}, ANI\footnote{\url{https://twitter.com/ANI}} for India and ABC\footnote{\url{https://twitter.com/ABC}}, CNN\footnote{\url{https://twitter.com/CNN}} for US based on accessibility, popularity and posts per day. Moreover, these Twitter handles are eminent for their objective and disinterested approach. The dataset has a total of 50000 samples, and each of the five categories has equal samples. The dataset has a Train-Val-Test split of 70:15:15.

\section{Related Work}
\subsection{Fake News Detection}
Recently, researchers proposed many effective methods to detect fake news. Qian et al.\cite{qian2018neural}showed how text content and semantic information could lead to the detection of fake news. In \cite{bhattarai2021explainable}, the authors used a new framework TM to capture lexical and semantic properties of both true and fake news text for detecting fake news. Besides, some lawbreakers tamper with news by forging fake images. So Qi et al.\cite{qi2019exploiting}proposed a novel model MVNN which utilizes the visual information of frequency and pixel domains to detect fake news. 

Though all the above uni-modal techniques have achieved good results, they still ignore the truth that news events often contain both text and pictures, which can complement each other. This indicates that we need a multi-modal system for fake news detection. The existing fake news detection methods based on multi-modal information can be divided into three categories. Some methods combined textual information with visual information. Singhal et al.\cite{singhal2019spotfake}  proposed a new model SpotFake to concatenate visual features extracted from the VGG-19 model pre-trained on ImageNet with textual features and classified using a fully connected layer. Wang et al.\cite{wang2018eann}built an end-to-end model termed EANN(event adversarial neural network)  to learn and concatenate the textual and visual latent representations and then feed them into two fully connected neural network classifiers. Other works focus on contrasting multi-modal information.SAFE\cite{zhou2020safe}, a similarity-aware multi-modal method, compared the relevance between the extracted features across modalities to detect fake news. In addition, some methods utilized multi-modal information enhancement to predict fake news. Jin et al.\cite{jin2017multimodal}used Recurrent Neural Network with an attention mechanism to enhance information understanding between multi-modal features for effective detection.

\subsection{Fact Verification}
In general, fact verification is the task of assessing the authenticity of a claim supported by a validated corpus of documents. Fact-checking introduces objective references and external knowledge to fake news detection, which is more reliable than relying simply on news text and visual features to make judgments. Fact verification can also increase users’ awareness of precaution, which helps them try to find fact-checking information when exposed to fake news.

There are many types of fact-verified datasets, and many of them are uni-modal datasets. For example, Thorne et al. \cite{thorne2018fever} considered Wikipedia as the source of textual evidence and annotated the sentences that support or refute each claim. Another popular type of unstructured evidence dataset often considered is metadata which offers information complementary to textual sources \cite{potthast2017stylometric}. Although most uni-modal datasets focus on unstructured evidence, structured knowledge has also been used. Some datasets consist of semi-structured data tables with the ability to convey information concisely and flexibly. Wang et al.\cite{wang2021semeval} extracted tables from scientific articles and required evidence selection in the form of cells selected from tables.

Recently, fact verification has also begun to consider building multi-modal datasets to improve the accuracy of fake news detection. A dataset named Mocheg\cite{yao2022end} consisted of textual and visual evidence and used three labels: support, refute, and NEI. Many fact verification methods focus on claim verification, which can be seen as natural language inference tasks or a form of Recognizing Textual Entailment(RTE). Nie et al.\cite{nie2019combining}used ESIM to verify a claim by concatenating all pieces of evidence as input and using the max pooling to aggregate the information. Typical retrieval strategies of RTE include search APIs, Lucene indices etc \cite{thorne2018fact}. In \cite{maillard2021multi}, dense retrievers using learned representations and fast dot-product indexing have performed well. Fan et al. \cite{fan2020generating} proposed another way to retrieve evidence by using question generation and question answering via search engine results.

\section{Methodology}
In our paper, we design a structure coherence-based fact-checking method in which the structure coherence between claims and documents is computed. Structure coherence is reflected in the following four aspects: literal text similarity, text semantic similarity, text length, and image similarity. Specifically, the ROUGE\cite{lin2004rouge} is used to extract the literal similarity of the text, two pre-trained models of CLIP\cite{reimers2019sentence} and Sentence BERT\cite{gao2021logically} are used to compute the semantic similarity of the text, the length of the claims and documents are calculated. Then ResNet50 is used to extract the image features. The above features are spliced and input into the random forest classifier to obtain the final classification result.

\begin{figure}
  \centering
  \includegraphics[width=\linewidth]{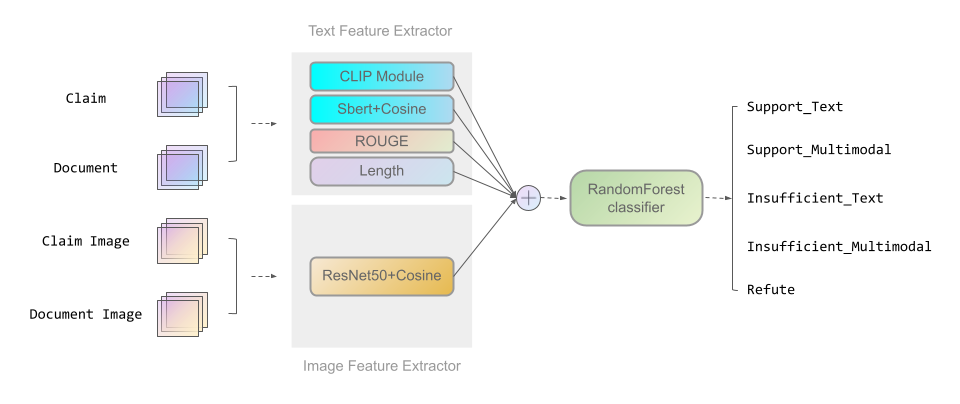}
  \caption{The overall architecture of our approach. It includes two parts: text feature extraction and image feature extraction. The CLIP module in text feature extraction is described in detail in section 4.}
  \label{fig:2}
\end{figure}

The architecture of the multi-modal model is shown in Figure \ref{fig:2}.

\subsection{Text Feature Extractor}
In the feature extraction module, two features of ROUGE and text length are extracted. ROUGE stands for Recall-Oriented Understudy for Gisting Evaluation. It is an evaluation metric that includes measures to automatically determine the quality of a summary by comparing it to other (ideal) summaries created by humans. The measures count the number of overlapping units such as n-grams, word sequences, and word pairs between the computer-generated summary to be evaluated and the ideal summaries created by humans. Since we observe a lot of vocabulary overlap in the category of refute, ROUGE, which counts the number of overlapping units, is well suited for our task. At the same time, inspired by Gao et al.\cite{mishra2022factify}, we also calculated the length of the text and found a specific correlation between the text length and the category, so we also extracted the text length feature.

In the pre-training model module, we use the Sentence BERT and cosine similarity in the baseline[25]. In addition, we tried pre-training models CLIP, SimCSE\cite{gao2021simcse}, and RoBERTa\cite{liu2019roberta} to extract text features.CLIP is a two-stream model that passes text and vision through the transformer encoder and calculates the similarity of different image-text pairs through linear projection. At the same time, it uses contrastive learning to convert image classification into image-text matching tasks. CLIP uses 400 million pairs of graphic and text datasets from the network and uses text as image labels for training. SimCSE is a simple framework for comparing sentence vector representations, which shows good performance in sentence embedding tasks. RoBERTa is an enhanced version of BERT and a fine-tuned version of the BERT model improves BERT\cite{devlin2018bert} in terms of model size and training methods. In this part, we tried to use CLIP text encoder, SimCSE and RoBERTa to extract text features and use cosine similarity to calculate semantic similarity. We also pass the features extracted by the clip through the MLP layer and output the three-category result. Finally, we found that using the three-category result of CLIP combined with MLP as a feature is more conducive to improving the final F1 score.

\subsection{Image Feature Extractor}
In this part, we use the pre-trained ResNet50\cite{kasnesis2021transformer} and cosine similarity. At the same time, two other solutions were also tried, using CLIP to extract image features and calculate cosine similarity; inputting CLIP features into MLP to output three-category results.

\subsection{Classifier}
Finally, the extracted features are then concatenated and passed through a Random Forest classifier to output five-category results. A Random Forest is a meta-estimator that fits several decision tree classifiers on various sub-samples of the dataset and uses averaging to improve the predictive accuracy and control over-fitting. Among the classifiers provided in the baseline, we selected the Random Forest classifier that performed best on our features.

\section{Experiments and results}

\subsection{Experiment Setting}
To evaluate the performance of our baseline solutions, we use weighted average F1 for benchmarking on the validation set. All the experiments are implemented on a single NVIDIA Tesla T4 GPU with up to 15 GiB RAM. For the CLIP module, we use the pre-trained text encoder and only train the following MLP layer. MLP contains one hidden layer with 100 nodes and Adam\cite{2014Adam} optimizer. Because the order of magnitude of text length and other features varies greatly, we normalize the final features. The final Random Forest classifier with a max\_depth value of 40 performs best.

\subsection{Results}
Our model achieved the F1 score of 0.8079 on the test set, and the competition leaderboard is shown in Table\ref{tab:1}. Our method achieves 2nd place in this competition. 

\begin{table*}\centering
  \caption{Factify Official Leaderboard}
  \label{tab:1}
  \begin{adjustbox}{center}
  \begin{tabular}{ccccccccl}
    \toprule
    Rank& Team& Support\_&  Support\_& Insufficient\_& Insufficient\_& Refute & Final\\
    & & Text&  Multi& Text& Multi&  & \\
    \midrule
    1& Triple-Check& 0.8277& 0.9138& 0.8518& 0.8922& 1& 0.8182\\
    \textbf{2}& \textbf{INO}& \textbf{0.8124}& \textbf{0.9003}& \textbf{0.8881}& \textbf{0.8523}& \textbf{0.9993}& \textbf{0.8080}\\
    3& Logically& 0.8038& 0.9051& 0.8439& 0.8563& 0.9851& 0.7897\\
    4& zhang& 0.7664& 0.8785& 0.8161& 0.8783& 0.9993& 0.7742\\
    5& gzw& 0.7849& 0.8632& 0.8142& 0.8327& 1& 0.7605\\
    6& coco& 0.7725& 0.8649& 0.8152& 0.8300& 1& 0.7570\\
    7& Noir& 0.7710& 0.8726& 0.7849& 0.8156& 0.9970& 0.7452\\
    8& Yet& 0.7075& 0.8263& 0.7859& 0.7190& 1& 0.6909\\
    9& TeamX& 0.5822& 0.7091& 0.5366& 0.5556& 0.6979& 0.4562\\
    \midrule
    -& Baseline& 0.5& 0.8272& 0.8024& 0.7593& 0.9882& 0.6499\\
  
  \bottomrule
\end{tabular}
\end{adjustbox}
\end{table*}

In the process of exploring solutions, our attempts are roughly divided into the following two aspects: 1) the selection of text pre-training models 2) the ways of using the CLIP.

During the experiment, we tried to use SimCSE, RoBERTa, and the text encoder of CLIP to replace Sentence BERT. Specifically, we use these pre-trained models as text feature extractors and then use cosine similarity to calculate the similarity between claim and document embeddings. The experimental results of only replacing Sentence BERT were compared with the baseline. The F1 score on the verification set is shown in Table \ref{tab:2}.

 It can be seen that using the text pre-training model to extract text features and simply calculate the cosine similarity, the F1 scores of other methods are lower than Sentence BERT. At the same time, we also tried using CLIP to extract image features and calculate cosine similarity to replace ResNet50. The result only reached an F1 score of 0.5103, far lower than the baseline. Therefore, simply replacing the pre-trained model without fine-tuning and computing the cosine similarity does not improve the classification performance. Because different pre-training models essentially only change the representation of the vector, no matter how large the dimension of the vector is, only one score will be obtained after calculating the cosine similarity. However, the information contained in a score is limited. So we still keep the structure of Sentence BERT and ResNet50.

\begin{table*}
  \caption{Replace Sentence BERT with other pretrained model}
  \label{tab:2}
  \begin{tabular}{ccl}
    \toprule
    Model Name & Validation F1 score\\
    \midrule
    SimCSE+ResNet50 & 0.6442\\
    RoBERTa+ResNet50 & 0.6424\\
    CLIP\_text+ResNet50 & 0.6404\\
    CLIP\_text+CLIP\_img & 0.5103\\
    Sentence BERT+ResNet50(baseline) & 0.6664\\
  \bottomrule
\end{tabular}
\end{table*}

\begin{figure}
  \centering
  \includegraphics[width=\linewidth]{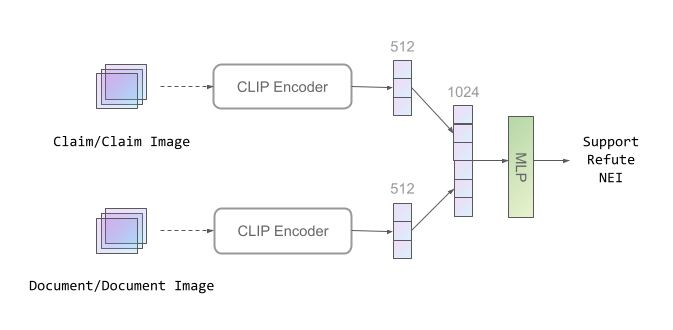}
  \caption{Method 1/2 for using CLIP in domain. Specifically, the text/image encoder of CLIP is used to extract the embedding of claim and document/claim\_image and document\_image, and input into MLP for three-category.}
  \label{fig:3}
\end{figure}

\begin{figure}
  \centering
  \includegraphics[width=\linewidth]{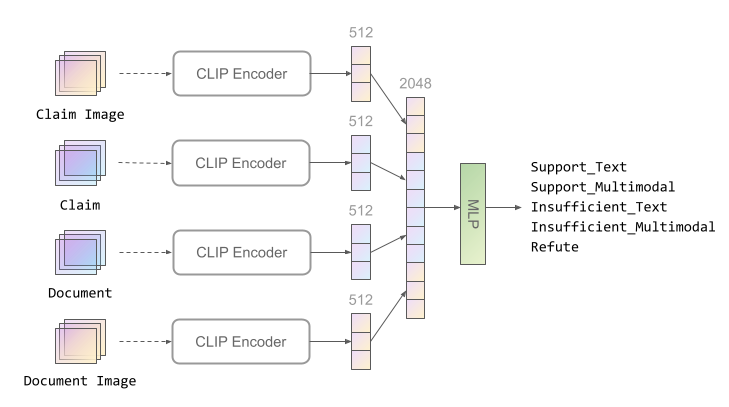}
  \caption{The third CLIP module, this method directly concatenates all the embeddings extracted by CLIP, and trains MLP to output five classification results.}
   \label{fig:4}
\end{figure}

In addition, we experimented with different ways of using CLIP. Because CLIP is a multimodal model, it contains a text encoder and an image encoder, both of which output a 512-dimensional vector. Therefore, after using CLIP to extract features from the data, we have a claim, document, claim image, doc image, and four 512-dimensional vectors. Therefore, we tried three feature combination methods to use the CLIP module, hoping that it can be used as a feature to improve the final classification effect.1)Concat the text feature vector into the MLP layer for three-category;2)Concat the image feature vector into the MLP layer for three-category;3)Concat all the image and text feature vectors, and input them into the MLP layer for five-category. The first two are shown in Figure \ref{fig:3}, and the third is shown in Figure \ref{fig:4}. The F1 scores on the validation set using different clip modules in our model are shown in Table \ref{tab:3}.

It can be seen that the different ways of using the clip module have a more significant impact on the classification results. Other usage methods will not improve the effect of the model or even reduce the effect. Because the feature obtained by this module is actually just a multi-class label. Therefore, the inaccuracy of classification may affect other features. Methods that only use CLIP to extract image features have the worst results. We consider that classification of images by CLIP and ResNet50 are mutually exclusive, so adding this feature will reduce the effect.

\begin{table*}
  \caption{Different clip modules}
  \label{tab:3}
  \begin{tabular}{ccl}
    \toprule
    CLIP Module & Validation F1 score\\
    \midrule
    CLIP+MLP & 0.7330\\
    CLIP\_text+MLP+CLIP\_image+MLP & 0.7291\\
    CLIP\_image+MLP & 0.7092\\
    CLIP\_text+MLP(ours) & 0.8078\\
  \bottomrule
\end{tabular}
\end{table*}

\begin{figure}[htbp]
\centering
\subfigure[on the validation set]{\includegraphics[width=0.45\textwidth]{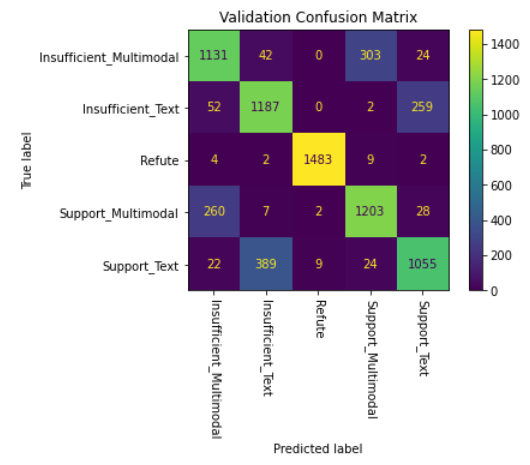}}
\subfigure[on the testing set]{\includegraphics[width=0.45\textwidth]{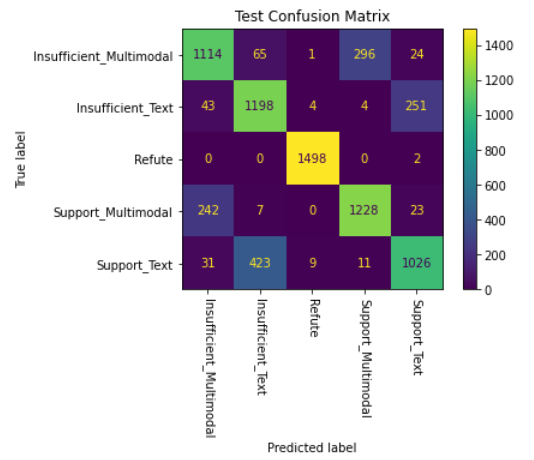}}
\caption{Confusion matrix of the validation set and testing set.}
\label{fig:5}
\end{figure}

Finally, we conducted ablation experiments on different modules in the model, and the results are shown in Table \ref{tab:4}. It can be seen that removing the  ResNet50 module has the most significant impact on the model effect. This result is expected since ResNet50 features are the only image features in the model. In addition, the most significant contribution to the model is the ROUGE and text length feature. At the same time, removing any text feature will not significantly impact the results.Figure\ref{fig:5} shows the confusion matrix of the final results on the validation and test sets.

\begin{table*}
  \caption{Ablation Study}
  \label{tab:4}
  \begin{tabular}{ccl}
    \toprule
    Model Name & Validation F1 score\\
    \midrule
    Without Sentence BERT & 0.7926\\
    Without CLIP & 0.7911\\
    Without ROUGE+length & 0.7709\\
    Without ResNet50 & 0.6007\\
    Baseline & 0.6664\\
    Ours & 0.8078\\
  \bottomrule
\end{tabular}
\end{table*}

\section{Conclusion}
This paper introduces a structure coherence-based approach for the multi-modal fact verification task. We tried different pre-training models such as SimCSE, CLIP, RoBERTa, and different ways of using CLIP multi-modal features. Finally, combining Sentence BERT and CLIP to extract text features and ResNet50 to extract image features achieved the best F1 score.

The future work can be carried out in four directions:1)Improve data preprocessing to address issues such as data noise and multilingualism;2)Our method does not fully use image features, consider introducing other features to complement ResNet50, or try other methods in the future;3)Explore better ways to use CLIP or adopt other methods to enhance modal fusion;4)Improve the universality of the fake news detection model and extend it to more types of datasets.

\section{Acknowledgement}

The work is supported by the National Key Research and Development Program of China (No.2020YFB1406800), the Fundamental Research Funds for the Central Universities (No.CUC22WH002), and the National Natural Science Foundation of China (61702502). We thank the organizers of DE-FACTIFY 2023 for allowing us to work on the dataset. We also thank Google colab for providing GPU and deep learning code running platform, and thanks to skit-learn and hugging face for providing an efficient and convenient deep learning library.

\bibliography{INO}
\end{document}